\documentclass[runningheads]{llncs}
\usepackage{tabularx,multirow,multicol}
\usepackage{graphicx, verbatim}
\usepackage{svg}
\usepackage[utf8]{inputenc}
\usepackage[T1]{fontenc}
\usepackage{amsfonts}
\usepackage{amsmath}
\usepackage{pifont}

\usepackage{hyperref}
\usepackage{color}

\usepackage{xcolor}
\usepackage{soulutf8}

\sethlcolor{cyan}

\newif\ifshowchanges
\showchangesfalse 

\newcommand{\change}[1]{%
 \ifshowchanges
   \hl{#1}%
 \else
   #1%
 \fi
}

\begin{document}
\title{CAT-SG: A Large Dynamic Scene Graph Dataset for Fine-Grained Understanding of Cataract Surgery}
\titlerunning{CAT-SG}
%
\author{\change{Felix Holm}\inst{1, 2}\orcidID{0000-0002-8493-9941}  \and
\change{Gözde Ünver}\inst{1, 2} \orcidID{0009-0000-5219-8926} \and \\
\change{Ghazal Ghazaei}\inst{2}\orcidID{0000-0002-0792-3583} \and
\change{Nassir Navab}\inst{1}}
\authorrunning{F. Holm et al.}
%
\institute{\change{Chair for Computer-Aided Medical Procedures, Technical University Munich, Germany} \and
\change{Carl Zeiss AG, Munich, Germany}\\
\email{\change{felix.holm@tum.de}}}


\maketitle              
\begin{abstract}
Understanding the intricate workflows of cataract surgery requires modeling complex interactions between surgical tools, anatomical structures, and procedural techniques. Existing datasets primarily address isolated aspects of surgical analysis, such as tool detection or phase segmentation, but lack comprehensive representations that capture the semantic relationships between entities over time. This paper introduces the Cataract Surgery Scene Graph (CAT-SG) dataset, the first to provide structured annotations of tool-tissue interactions, procedural variations, and temporal dependencies. By incorporating detailed semantic relations, CAT-SG offers a holistic view of surgical workflows, enabling more accurate recognition of surgical phases and techniques. Additionally, we present a novel scene graph generation model, CatSGG, which outperforms current methods in generating structured surgical representations. The CAT-SG dataset is designed to enhance AI-driven surgical training, real-time decision support, and workflow analysis, paving the way for more intelligent, context-aware systems in clinical practice.

\keywords{Scene Graphs  \and Surgical Data Science \and Dataset \and Cataract Surgery}

\end{abstract}
\section{Introduction}

Advancements in machine learning and computer vision are transforming com-puter-assisted surgery, enabling precise planning, real-time decision support, and automated workflow analysis. While progress has been made in surgical video analysis for tasks like tool detection, phase segmentation, and triplet recognition, achieving a holistic yet fine-grained understanding of surgical workflows remains a challenge.

Scene graphs offer a robust framework for modeling structured interactions within complex environments, representing entities as nodes and their relationships as edges. In computer vision research, they are increasingly recognized as valuable tools for providing structured, fine-grained, and human-readable representations that yield holistic insights into dynamic scenes \cite{ji2020action,moma,ego4d}. Therefore,  the application of dynamic scene graph representations holds significant promise for revolutionizing surgical workflow analysis by encoding surgical scenes through surgical elements, their spatio-temporal features, and interactions. Although prior research has explored scene graph-based approaches for surgical workflow recognition \cite{murali,holm2023dynamic,sangria}, these efforts have been constrained by the limitations of existing datasets lacking fine-grained semantic interactions necessary for detailed procedural analysis. 

Cataract surgery, one of the most common and intricate ophthalmic procedures, exemplifies the need for structured surgical representations. The success of phacoemulsification, a widely used cataract removal technique, depends on precise tool-tissue interactions, such as controlled ultrasonic energy application for lens fragmentation and delicate intraocular manipulation. Cataract procedures involve fine-grained motion dynamics and subtle anatomical interactions, which existing datasets fail to systematically capture, hindering the development of automated workflow analysis solutions. 

To address this gap, we introduce the Cataract Surgery Scene Graph (CAT-SG) dataset, a novel resource derived from the CATARACTS dataset~\cite{CATARACTS}. CAT-SG provides structured representations of surgical elements, encompassing their geometric, temporal, and semantic interactions, with a particular focus on tool-tissue relationships. It also introduces a new downstream task of surgical technique recognition, complementing the existing surgical phase annotations in CATARACTS.

Our contributions are summarized as follows: \\
\textbf{CAT-SG Dataset:} We present the first Cataract Surgery Scene Graph dataset, with over 1.8 Million annotated relations. Additionally, CAT-SG provides novel annotations for surgical techniques, enabling a more comprehensive understanding of procedural variations beyond traditional phase recognition. \\
\textbf{Comprehensive Benchmarking and Evaluation:} To demonstrate the utility of CAT-SG, we establish benchmarks for the three key tasks of our dataset: scene graph generation, phase recognition and technique recognition, demonstrating the dataset’s applicability across various surgical workflow tasks. \\
\textbf{Novel Scene Graph Generation Model (CatSGG):} Additionally, we propose CatSGG, a new scene graph generation model leveraging large-scale pre-training and spatio-temporal attention to efficiently generate structured representations of surgical interactions outperforming the state-of-the-art method ORacle~\cite{oracle} on CAT-SG.

\begin{figure}
  \centering
  \includegraphics[width=1\linewidth]{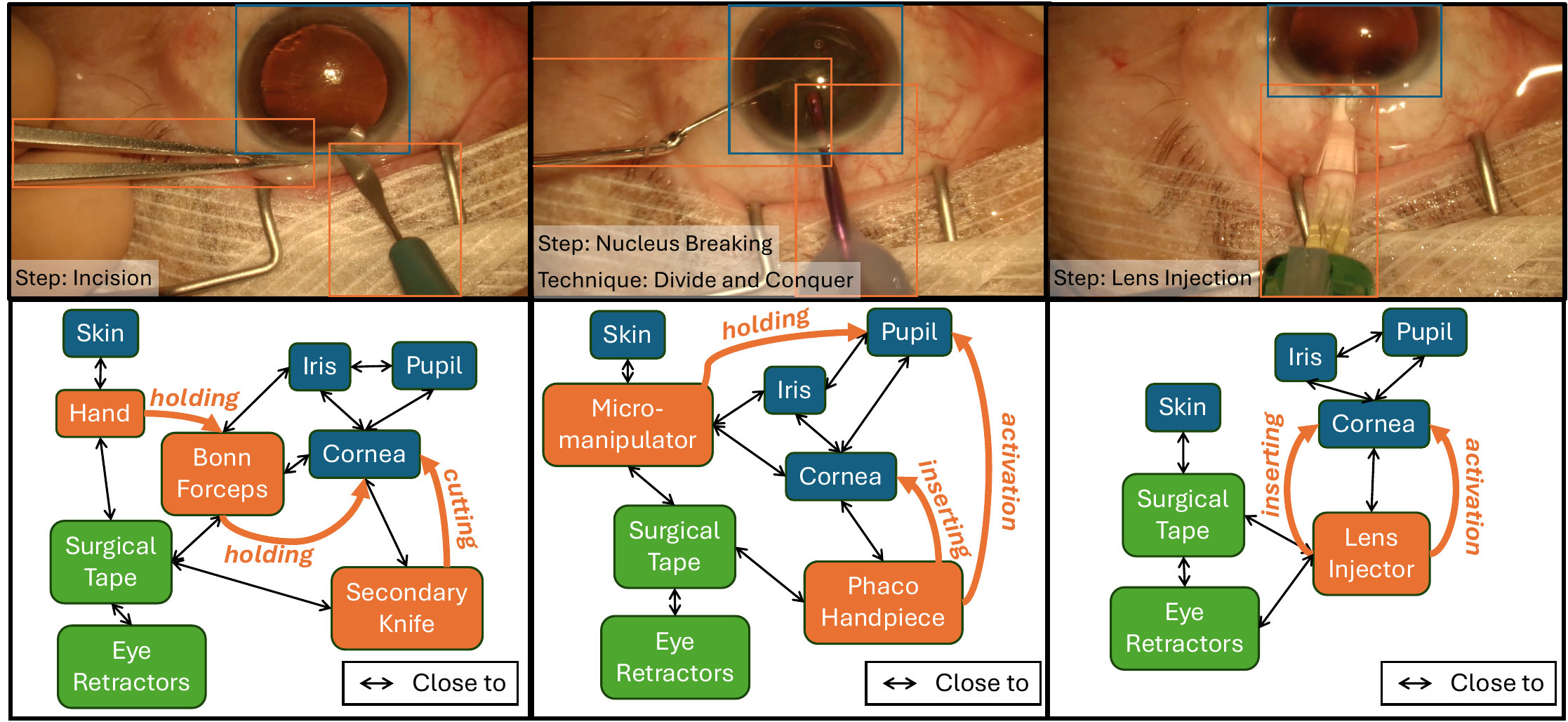}
  \caption{Samples from one video of CAT-SG, showing Image and Scene Graph including semantic and geometric relations}
  \label{fig:data_samples}
\end{figure}

\section{CAT-SG Dataset}

\subsection{CATARACTS}
We introduce a novel dataset named CAT-SG that extends the publicly available CATARACTS dataset~\cite{CATARACTS} with scene graph annotations. The CATARACTS dataset consists of 50 high-resolution videos of cataract surgeries, collected at Brest University Hospital. Each video captures a complete surgical procedure with a duration of 11 minutes on average (min: 6 min, max: 40 min). With over nine hours of annotated surgical footage, CATARACTS provides a robust basis for developing machine learning models for surgical video understanding~\cite{CATARACTS}.

\subsection{Annotations}
The scene graph annotations were created through a manual labeling process conducted by 9 trained student annotators. The annotators reviewed the videos and interactively marked each instance of a surgical tool interacting with an anatomical structure\change{, followed by an iterative expert review process to verify the validity of our labels}. After over 1200 combined hours of annotating, this process resulted in a detailed temporal mapping of tool-tissue interactions throughout the surgeries. The interactions are classified into 8 categories shown in Table~\ref{tab:relations}.

\begin{table} [b!]
    \caption{Relations annotated in CAT-SG}
\fontsize{9pt}{9pt}\selectfont
    \centering
    \begin{tabular}{l|p{8cm}|r}
 Relation& Example&Samples\\ \hline
        Holding & e.g., forceps fixating sclera& 13,380\\ \hline
        Activation & e.g., phaco-handpiece using emulsification & 44,552\\ \hline
 Pushing& e.g., micro manipulator pushing \& rotating nucleus&3,874\\ \hline
 Pulling& e.g., cystotome pulling on capsular bag flap&11,895\\ \hline
 Cutting& e.g.,  corneal incisions using knife&1,925\\ \hline
 Inserting& \multirow{2}{8cm}{tool is inserted or retracted through the primary or secondary incision in the cornea} &34,016\\
 Retracting& &23,886\\ \hline
 Close to& spatial proximity of one object to another&1,677,724\\
    \end{tabular}
    \label{tab:relations}
\end{table}

The following 29 objects were considered for annotations: Pupil, Surgical Tape, Hand, Eye Retractors, Iris,
Skin, Cornea, Hydrodissection Cannula, Viscoelastic Cannula, Capsulorhexis Cystotome, Rycroft Cannula, Bonn Forceps, Primary Knife, Phacoemulsification Handpiece, Lens Injector, Irrigation/Aspi-ration Handpiece,
Secondary Knife, Micromanipulator, Capsulorhexis Forceps, Suture Needle, Needle Holder, Charleux Cannula, Vitrectomy Handpiece, 
Mendez Ring, Marker, Troutman Forceps, Cotton, Iris Hooks, Vannas Scissors.

We sample and annotate the 50 videos in our dataset at a temporal resolution of 5 fps, making CAT-SG the largest dataset among comparable efforts for detailed surgical scene understanding (Table \ref{tab:datasets}). We also use pseudo-labelled segmentation masks based on the CaDIS dataset to provide a grounding to each object in our annotations (position, size, bounding box). We will also release these masks in addition to our dataset.
\begin{table} [t]
    \caption{Size of comparable datasets and CAT-SG}
\fontsize{9pt}{9pt}\selectfont
    \centering
    \begin{tabular}{c|cccccc}
    && Avg. &  Annotated &  Unique &  Unique & Annotated \\
         Dataset&  Videos&   Duration& Frames&  Objects&  Relations & Relations\\ \hline
         4D-OR&  10&  11 min&  6,743&  11&  14& 103,740\\
         CholecT45&  45&  33 min&  90,489&  20&  9& 127,385\\
         CAT-SG&  50&  11 min&  164,162&  29&  9& 1,811,252\\ \hline
    \end{tabular}
    \label{tab:datasets}
\end{table}
%
%
\subsubsection{Use Cases:}
CAT-SG is designed to support detailed surgical video understanding through structured scene graph representations offering the following tasks:\\
\textbf{Surgical Scene Graph Generation:} This task involves automatically constructing structured scene graphs from surgical videos. Given a sequence of frames, the goal is to detect surgical instruments, anatomical structures, and their interactions over time. Successful models must not only recognize objects but also infer their relationships, enabling a deeper understanding of surgical workflow and tool usage patterns.\\
\textbf{Surgical Workflow Recognition:} Our dataset enables surgical workflow recognition by leveraging the predefined 19 surgical steps from the CATARACTS dataset~\cite{CATARACTS}. With the integration of scene graph annotations, this process gains additional structure and granularity, reinforcing its role as a fundamental task in surgical video understanding.\\
\textbf{Surgical Technique Recognition:} To enhance surgical video analysis, we add new annotations distinguishing two nucleus-breaking techniques: “Stop and Chop” and “Divide and Conquer.” \change{Each procedure in the dataset is labeled as using one of these two techniques during the nucleus breaking step.} Identifying these techniques requires a nuanced understanding of tool movements and procedural dynamics, making this a complex and valuable \change{benchmark} for automated surgical workflow analysis.
\section{CatSGG Scene Graph Generation Method}
\label{sec:methods}

We introduce a novel scene graph generation method comprising of two steps: 1) entity localization and base graph construction, 2) geometric and semantic relation prediction. \\
\begin{figure} [t!]
  \centering
  \includegraphics[width=1\linewidth]{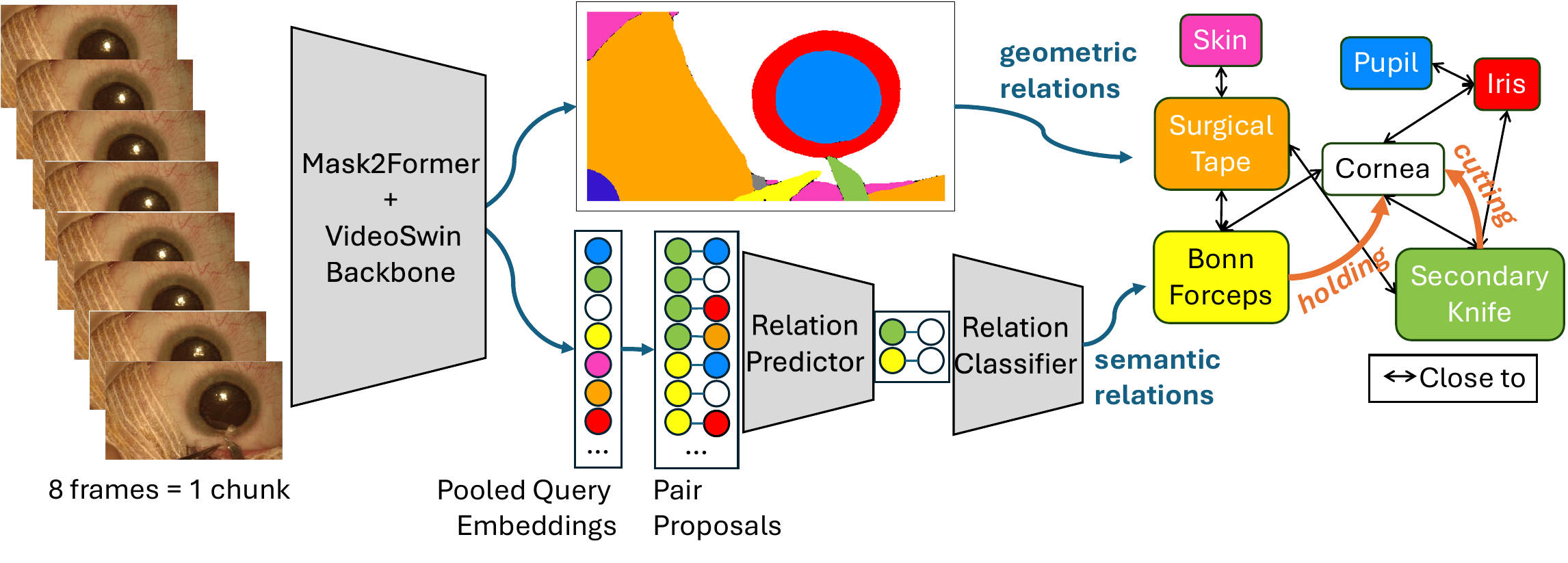}
  \caption{Overview of the CatSGG+ Pipeline: an input video chunk is fed into Mask2Former with a video-based backbone to extract rich spatio-temporal features. The extracted query embeddings are pooled and used to generate pair proposals, which are then processed by a relation predictor and classifier to infer semantic relations. Geometric relations are inferred from the predicted mask.}
  \label{fig:CatSGG-pipeline}
\end{figure}

\textbf{Annotation-efficient Entity Localization} 
While bounding boxes provide coarse localization, they lack the spatial precision needed for modeling surgical interactions in scene graphs. We leverage segmentation models instead, as they offer fine-grained instance localization, enabling more accurate spatial reasoning and relationship extraction. Surgical video segmentation could also be challenging due to the long duration of procedures, e.g. cataract surgeries typically last around 11 minutes, making manual annotation and computational processing at high fps highly burdensome. To address this, we propose an annotation-efficient solution for instance segmentation of surgical elements within CATARACTS frames by leveraging state-of-the-art segmentation and video-language pretraining.  
Our approach builds upon Mask2Former (M2F) \cite{cheng2022masked} while incorporating domain-specific prior knowledge from Watch\&Learn \cite{watchandlearn}, a large-scale video-language pretraining framework. Specifically, we replace the standard backbone of M2F with a finetuned VideoSwin \cite{liu2022video}, which was trained within the VALOR framework \cite{chen2023valor} alongside a BERT encoder \cite{devlin2019bert} for video-language alignment. We opt for this model over the original VideoSwin because its pretraining on 2,900 YouTube cataract surgery videos \cite{watchandlearn} provides strong domain priors, allowing effective segmentation with fewer annotated samples. Moreover, the spatio-temporal design of VideoSwin enhances the model’s awareness of cataract surgery dynamics, further improving segmentation performance.  
By integrating VideoSwin into M2F, we adapt the segmentation pipeline to better capture the temporal and contextual dependencies in surgical videos while requiring fewer annotated frames.\\
%
\textbf{Geometric Relation Prediction:} The “Close to” relation, which captures spatial proximity, is determined based on M2F segmentation predictions. Specifically, we identify instances with adjacent \change{masks (touching boundaries)} as being close to each other, following the approach in \cite{holm2023dynamic}.\\
\textbf{Semantic Relation Prediction:} \label{sem-rel-pred}
The VideoSwin backbone of our M2F processes 8 consecutive frames as a chunk, with segmentation and scene graph generation performed only on the last frame. Consequently, our backbone inherently captures temporal information across frames.
 Inspired by query embedding solutions \cite{wang2024oed,wang2024pair,yang2023panoptic,zhu2025towards,zhou2023hilo,yang2022panoptic}, we leverage the output query vectors ($q$) from M2F for relation prediction. Each query vector encodes class-specific and spatial information for a single instance. 
 To model pairwise relationships, we construct pair proposal embeddings by concatenating the queries of two instances. Since semantic relations occur only between tool-tool or tool-anatomy pairs, we incorporate dataset priors to construct these pair proposals efficiently.
 Relation prediction consists of two tasks: (1) detecting whether a relation exists ($e\geq0.5$), and (2) classifying the relation type. Following \cite{yang2023panoptic}, we employ: A 2-layer model for binary relation existence prediction ($f_{\text{existence}}$), followed by a 3-layer multi-label classification model ($f_{\text{classification}}$) for relation type prediction.

 \begin{equation} \label{eq:mlp_equation}
 \begin{split}
 \text{pair}_n&=[q_i;q_j] \textit{ where } i \neq j,i \in N_{\text{tool}},j\in N_{\text{tool}}\cup N_{\text{anatomy}} \\
 e&=\sigma(f_{\text{existence}}(\text{pair}_n)), e \in [0,1] \\
 \textit{if } e\geq0.5,c&=f_{\text{classification}}(\text{pair}_n)
 \end{split}
  \end{equation}
We use binary cross-entropy loss with sigmoid activations for both models. 

Both models operate on the pair embeddings derived from the query vectors of the last frame in the chunk. 
We refer to this method as CatSGG.
To further enhance temporal consistency, we extend this approach by aggregating query vectors from all frames in the chunk. Inspired from \cite{yang2023panoptic}, we apply max pooling on same-class queries across these frames to obtain a single representative vector per class.

These temporally enriched vectors are then used to create pair embeddings for relation predictions. We refer to this temporally aware extension as CatSGG+.

\section{Experiments}

\textbf{CatSGG: Semantic Segmentation:} \label{Sg-ins-seg}
From preliminary experiments we find that training M2F with the pretrained VideoSwin backbone from\cite{watchandlearn} achieves equal performance with much faster convergence and improved generalizability by subsampling the training data to 18 chunks per training video.
Using this configuration 
we train M2F on the 29 classes from CAT-SG, by merging pseudo-masks based on CaDIS \cite{grammatikopoulou2019cadis} with CAT-SG annotations. 
The model achieves a mIoU of 92.12\%, demonstrating its ability to generate initial scene graph nodes.\\
\textbf{CatSGG: Semantic Relation Prediction:}
Using the frozen pretrained segmentation model, we train the relationship prediction models with 18 chunks per training video. Since semantic relations are relatively infrequent, we apply sampling constraints to ensure that each selected chunk contains at least one semantic relation. We apply a 0.5 threshold to filter predictions from both the relation existence and classification models.\\

\textbf{ORacle \cite{oracle}:}
We use ORacle as a benchmark, representing the most recent SOTA method in scene graph generation within the surgical setting. ORacle is a large vision-language model and does not require bounding boxes or scene graph grounding, which makes it highly versatile and easily adaptable to various domains. ORacle is originally developed for the 4D-OR dataset, consisting of external views captured by overhead cameras in the operating room (OR). We train and evaluate ORacle’s single view variant, both with and without temporality (denoted as ORacleSV and ORacleSVT, respectively). To account for the distribution of relation classes, we report F1 scores for each class, as well as micro- and macro-averaged F1.

\subsubsection{Downstream Task Baselines:}
Since semantic scene graphs connect the graph and language domain, they can be used both in Graph Neural Networks (GNNs) and Large Language Models (LLMs). We present baselines for both approaches, leveraging our scene graphs to recognize surgical workflows and  techniques. We report Accuracy and F1 score as the well-established metrics for these tasks~\cite{funke2023metricsmattersurgicalphase}.\\
\textbf{GNN:} 
We use a 3-layer GATv2~\cite{gatv2} model as our GNN baseline. Like \cite{holm2023dynamic}, we encode node groundings (i.e., position and size in the image) as part of the node features. Temporality is captured by connecting scene graphs from different timesteps with temporal edges between nodes of the same class, creating a dynamic scene graph.\\
\textbf{LLM:} 
We fine-tune a Llama 3.2 3B model for our downstream tasks using QLora and its recommended parameters~\cite{dettmers2023qlora}. To represent our scene graphs, we prompt the model with a list of entities and relations, along with their groundings provided as text (e.g., “<Object> at (<x>, <y>) with size <size>”). Temporality is incorporated through a history mechanism: the prompts include scene graphs from the last N steps prior to the current scene graph that needs to be classified.

\section{Results \& Discussion}
\textbf{Scene Graph Generation.} 
Table \ref{tab:results_sgg} presents the results for scene graph generation on CAT-SG. Our CatSGG outperforms the SOTA ORacle by over 8 percentage points (p.p.) on Macro-F1. 
The inclusion of longer-term temporal context in CatSGG+ further improves performance, allowing the model to better capture the dynamically evolving semantic relations over time. Surprisingly, adding temporality to ORacle (ORacleSVT) leads to a performance drop. We hypothesize this is due to ORacle aggregating its predictions over time, which can propagate mispredictions, thereby degrading overall performance. A video of qualitative results for CatSGG+ is attached in the supplementary.

\begin{table}
    \caption{Scene Graph Generation Results on CAT-SG.}
\fontsize{9pt}{9pt}\selectfont
    \centering
    \begin{tabular}{l|ccccccccc|cc}
    & \multicolumn{9}{c}{F1 Score for each Relation} & \multicolumn{2}{c}{Overall} \\
    & \rotatebox{85}{Close to} & \rotatebox{85}{Holding} & \rotatebox{85}{Activation} & 
      \rotatebox{85}{Pushing} & \rotatebox{85}{Pulling} & \rotatebox{85}{Cutting} & 
      \rotatebox{85}{Inserting} & \rotatebox{85}{Retracting} & \rotatebox{85}{"none"} & 
      \rotatebox{85}{Micro F1} & \rotatebox{85}{Macro F1} \\ \hline

    ORacleSV & 67.03& 3.74& 41.76& 1.57& 33.08& 31.06& \underline{39.75}& 15.00& 78.83& 72.45& 34.65\\
    ORacleSVT & 57.64 & \textbf{19.82} & 35.93 & \textbf{15.58} & 32.99 & 14.74 & \textbf{42.16} & \underline{17.10} & 74.48 & 66.19 & 34.49 \\
    CatSGG&  \textbf{91.63} & 7.52 & \textbf{45.81} & \underline{4.32} & \textbf{43.09} & \textbf{49.23} & 37.74 & 7.26 & \textbf{92.10} & \textbf{89.78} & \underline{42.08} \\
    CatSGG+ & \textbf{91.63} & \underline{8.60} & \underline{44.90} &  0.00 & \underline{42.86} & \underline{46.04} & 39.32 & \textbf{22.55} & \underline{92.08} &\textbf{89.78} & \textbf{43.11} \\ \hline
    \end{tabular}
    \label{tab:results_sgg}
\end{table}


\textbf{Surgical Workflow Recognition.} As shown in Table \ref{tab:results_workflow}, introducing semantic relations leads to improvements in both accuracy and F1, surpassing the baseline from \cite{holm2023dynamic} by over 5~p.p. This demonstrates that explicitly encoding tool–anatomy relationships is beneficial for recognizing workflow stages that often hinge upon the correct sequence of instrument interactions. Notably, LLMs performed competitively in our experiments without temporality, but were not able to leverage temporal context or groundings for improved performance, despite being much larger models.

\begin{table}
    \caption{Surgical Workflow Recognition Results}
\fontsize{9pt}{9pt}\selectfont
    \centering
    \begin{tabular}{l|ccc|cc} 
        & Temporal & Semantic  &Spatial& & \\
        &  Window&  Relations &Encoding&  Accuracy& F1\\ \hline
         Holm et al. \cite{holm2023dynamic} &  1 frame &  \ding{55} CATARACTS  &\ding{51} &  65.56 & 52.24\\
         GATv2 & 1 frame & \ding{51} CAT-SG  &\ding{51} & \textbf{70.81} & \textbf{56.02} \\
         Llama 3.2 3B & 1 frame & \ding{51} CAT-SG  &\ding{55}& 67.70 & 53.98 \\
 Llama 3.2 3B & 1 frame&  \ding{51} CAT-SG&\ding{51} & 69.13 & 53.87 \\ \hline
         Holm et al. \cite{holm2023dynamic} & 30 frames (90 s) & \ding{55} CATARACTS  &\ding{51} & 73.77 & 64.93 \\
         GATv2 & 30 frames (90 s) & \ding{51} CAT-SG  &\ding{51} & \textbf{78.63} & \textbf{70.15} \\
 Llama 3.2 3B & 30 frames (90 s)&  \ding{51} CAT-SG&\ding{55}& 29.02 & 5.52 \\ \hline

    \end{tabular}
    \label{tab:results_workflow}
\end{table}

\textbf{Surgical Technique Recognition.}Table~\ref{tab:results_technique} shows that our GATv2 model, utilizing temporal graphs sampled at 5~fps, achieves the highest accuracy and F1 scores. Interestingly, longer temporal windows sampled with lower resolution (50~s at 1~fps) result in lower accuracy, suggesting that the fine-grained instrument motions crucial for distinguishing specific techniques occur within shorter time spans.
This highlights the value of our dataset being annotated at 5 fps. Furthermore, ablating spatial features degrades performance, confirming that precise tool positioning is vital for differentiating lens-breaking strategies.

\begin{table}[b]
    \caption{Surgical Technique Recognition Results}
\fontsize{9pt}{9pt}\selectfont
    \centering
    \begin{tabular}{l|cc|cc}
 & Temporal Window& Spatial& &\\
         &  (sampled fps)&  Features& Accuracy & F1\\ \hline
         GATv2 & 10 s (5 fps) & \ding{51} &  \textbf{68.75} $\pm$ 4.11& \textbf{48.40} $\pm$ 1.72\\
         GATv2 & 50 s (1 fps) & \ding{51} &  66.34 $\pm$ 1.38& 41.31 $\pm$ 1.83\\
         GATv2 & 10 s (5 fps) & \ding{55} &  64.48 $\pm$ 1.15& 44.52 $\pm$ 1.58\\ \hline
    \end{tabular}
    \label{tab:results_technique}
\end{table}

\clearpage
\section{Conclusion}
In this paper, we introduce CAT-SG, a large, fine-grained scene graph dataset specifically designed to understand complex workflows in cataract surgery. By capturing not only the presence of surgical instruments and anatomical structures but also their intricate interactions, CAT-SG provides an unprecedented level of detail for modeling real-world procedural variations. Alongside the dataset, we propose CatSGG, a novel scene graph generation model that leverages pre-trained video backbones and efficiently processes both spatial and temporal information.
The holistic view offered by CAT-SG has the potential to enhance a wide range of applications, from automated skill assessment and decision support to advanced surgical training. Furthermore, this structured representation of surgical procedures promotes more transparent and explainable approaches, as each relationship can be precisely interpreted in a clinical context. Ultimately, this can improve patient outcomes and support surgeons in delivering safer, more efficient care.


    

\begin{credits}
\subsubsection{\ackname} \change{The authors would like to thank Carl Zeiss AG for support of this research.}

\subsubsection{\discintname}
\change{The authors have no competing interests to declare that are relevant to the content of this article.}
\end{credits}

%
%

\bibliographystyle{splncs04}
\bibliography{mybibliography}

\end{document}